\documentclass{dexmate_report}

\usepackage{amsmath,amsfonts,bm}

\def\eqref#1{equation~\ref{#1}}

\def\1{\bm{1}}

\DeclareMathAlphabet{\mathsfit}{\encodingdefault}{\sfdefault}{m}{sl}
\SetMathAlphabet{\mathsfit}{bold}{\encodingdefault}{\sfdefault}{bx}{n}

\usepackage{makecell}
\newcommand{\mainSeedSpread}{3.4}
\newcommand{\appBootN}{1{,}000}
\newcommand{\appMainPool}{five backends (wiki, A-MEM, Mem0, ObsMem, and transcript)}
\newcommand{\appMainRuns}{795}
\newcommand{\appRunsPerTask}{fifteen}
\newcommand{\appWritePool}{wiki, A-MEM, Mem0, and ObsMem}

\newcommand{\benchname}{MeMoREval}
\graphicspath{{figs/}}

\title{What Should Embodied Agents Remember? Benchmarking and Evaluating Memory Modality for Embodied Agents}
\shorttitle{MeMoREval: Memory Modality for Embodied Agents}
\author{Junxiang~Wang\textsuperscript{1,2\textdagger}, Yuchen~Liu\textsuperscript{1,3\textdagger}, Yi~Wu\textsuperscript{1}, Guofei~Chen\textsuperscript{1}, Nir~Pechuk\textsuperscript{2},\\
Tao~Chen\textsuperscript{1}, Zackory~Erickson\textsuperscript{2}, Yuzhe~Qin\textsuperscript{1} \& Tianyu~Li\textsuperscript{1}}
\affiliation{\textsuperscript{1}Dexmate,\enspace\textsuperscript{2}Carnegie Mellon University,\enspace\textsuperscript{3}Purdue University}
\authornote{\textsuperscript{\textdagger}Work done during an internship at Dexmate.}
\date{Oct 2026}
\dexmateabstract{\begin{abstract}
Memory allows large language model (LLM) agents to use past experience, often stored as textual records, when responding to requests. For embodied agents that control robots, visual and spatial observations introduce a broader choice of memory modalities, affecting both the information available for subsequent tasks and the model inference costs of constructing and using memory. We introduce \benchname, a benchmark for studying this problem through a controlled experience-request protocol. Agents first \emph{experience} a sequence of events without knowing the subsequent tasks, and subsequently they use memory to respond to information and action \emph{requests} without receiving new observations. The benchmark comprises 53 tasks across four simulated indoor environments and measures task success alongside the inference costs of memory construction and use. Our reference evaluations reveal that storing frames outperforms storing generated descriptions at lower write-time inference cost for the primary model, while storing frames with scene graphs achieves the highest aggregate task success for every tested model. Extensions to accumulated experience, selective frame retention, and visual input to memory writers demonstrate how the benchmark accommodates further memory-design investigations. Two physical-robot demonstrations establish the feasibility of applying its protocol beyond simulation.
\end{abstract}
}
\reportlinks{%
  \reportlink{Code}{https://github.com/jimwang418/MeMoREval}%
  \reportlink{Website}{https://MeMoREval.github.io/}}
\correspondence{junxiang@cmu.edu; ty.li@dexmate.ai}
\hypersetup{
  pdftitle={What Should Embodied Agents Remember? Benchmarking and Evaluating Memory Modality for Embodied Agents},
  pdfauthor={Junxiang Wang, Yuchen Liu, Yi Wu, Guofei Chen, Nir Pechuk, Tao Chen, Zackory Erickson, Yuzhe Qin, Tianyu Li},
  pdfsubject={Dexmate Technical Report -- MeMoREval}}

\begin{document}
\maketitle

\section{Introduction}
\label{sec:intro}

A coding agent investigating a bug may uncover facts about a codebase that remain useful long after the immediate fix. Memory systems preserve such findings beyond the model's current context~\citep{mem0,amem,memgpt,reflexion,awm}. For large language model (LLM) agents that control robots, what is worth preserving extends to camera frames, object locations, and whether physical actions succeeded. This experience can be retained as textual descriptions, raw images, structured spatial records, or combinations of these forms, and the choice affects both what survives and what memory costs.

A textual record might note that several mugs were on a counter while omitting the colors needed to later retrieve a particular mug. Generating a description requires selecting what to mention before the future task is known, and task-relevant details may be discarded. A stored camera frame retains visual details for interpretation once the task is known, while a structured spatial record makes object locations and relations directly available.

Decisions to remember certain information are particularly important for embodied agents because recovering forgotten information can require considerable physical action. A coding agent can often rediscover a fact by inspecting an available codebase again or inspecting an earlier version of the codebase, which incurs additional computation and latency, but the memory lapse is largely invisible to the user. For a robot, recovering missing information may require navigating between rooms, opening containers, or manipulating objects. Such recovery makes memory failures apparent through repeated searches and unnecessary physical actions, delaying task completion. Moreover, changes to the environment can make forgotten information irrecoverable: once objects have been moved or removed, their earlier spatial relationships may no longer be observable.

To study which forms of retained evidence best support subsequent recall and action, we introduce \benchname{}, a controlled \textbf{Eval}uation protocol and benchmark for studying \textbf{Me}mory \textbf{Mo}dalities of \textbf{R}obotic embodied agents. We distinguish the \emph{memory modality}, which determines the form of retained evidence, from the \emph{memory backend}, which organizes that evidence and makes it available to the agent. Our study addresses three questions: (\textbf{RQ1}) How do memory modalities affect success on information and action requests across task categories and models? (\textbf{RQ2}) How do these modalities trade off task success against write-time and read-time inference cost? (\textbf{RQ3}) How can \benchname{} inform memory-design decisions for deploying embodied agents?

Our evaluation protocol constructs each task from two phases. In the experience phase, the robot follows a scripted sequence of events while the memory module records its observations. In the request phase, the agent starts from a fresh context and uses memory to answer information requests and plan action requests, as a robot would when a user poses a request. Plans are executed without returning new observations, so the agent cannot compensate for missing memory by searching again, which would reward exploration ability rather than memory. Experience phases can also be stacked to evaluate memory over longer histories.

We instantiate this protocol in 53 simulated tasks across four simulated indoor environments and seven categories of remembered information. \benchname{} supports textual records, generated image descriptions, camera frames, scene graphs, and their combinations. Using common recorded observations and a fixed agent implementation, we compare different memory configurations on task success as well as model inference costs, as both write-time and read-time costs are candidates to be optimized in embodied agent design.

Our reference evaluations show that storing camera frames with scene graphs achieves the highest aggregate task success for every tested model. For the primary model, archiving frames achieves higher success and lower write-time inference cost than generating descriptions, while also yielding the lowest read-time inference cost per successful request. The benefits of individual modalities vary across models and task categories, demonstrating the value of evaluating representations together with the models that use them.

We further examine deployment-oriented variants that accumulate experience across tasks, let writers select which frames to retain, or provide images to writers during memory construction. These variants preserve the benchmark's evaluation structure while extending the design choices under study. The resulting differences in task success, retention rates, and inference costs illustrate how \benchname{} can support investigations of memory organization, evidence selection, and writer inputs. Two demonstrations on a physical robot further establish that the protocol and memory interfaces can be instantiated using camera observations and an annotated spatial map.

We situate the memory-modality question within prior work on LLM and embodied agents and their evaluation (\S~\ref{sec:related}), then establish the agent model and memory terminology needed to study it (\S~\ref{sec:preliminary}). We then introduce the design of our evaluation protocol and its instantiation in \benchname{} (\S~\ref{sec:methods}). Our experiments address the three research questions and demonstrate the protocol on a physical robot (\S~\ref{sec:results}).

Our contributions are:
\begin{itemize}[topsep=0pt,partopsep=0pt,parsep=0pt,itemsep=0.1em,leftmargin=1.5em,labelsep=0.5em]
\item A controlled memory evaluation protocol for embodied agents that separates experience from subsequent information and action requests, supports accumulated histories, and measures write-time and read-time inference costs.
\item \benchname{}, a benchmark of 53 tasks across four simulated environments, with configurable memory modalities and backends and complementary demonstrations on a physical robot.
\item Reference evaluations across memory modalities, models, and backends that characterize trade-offs between task success and inference cost and identify opportunities for improving evidence retention and access in deployed embodied agents.
\end{itemize}

\begin{figure}[t]
  \includegraphics[width=\textwidth]{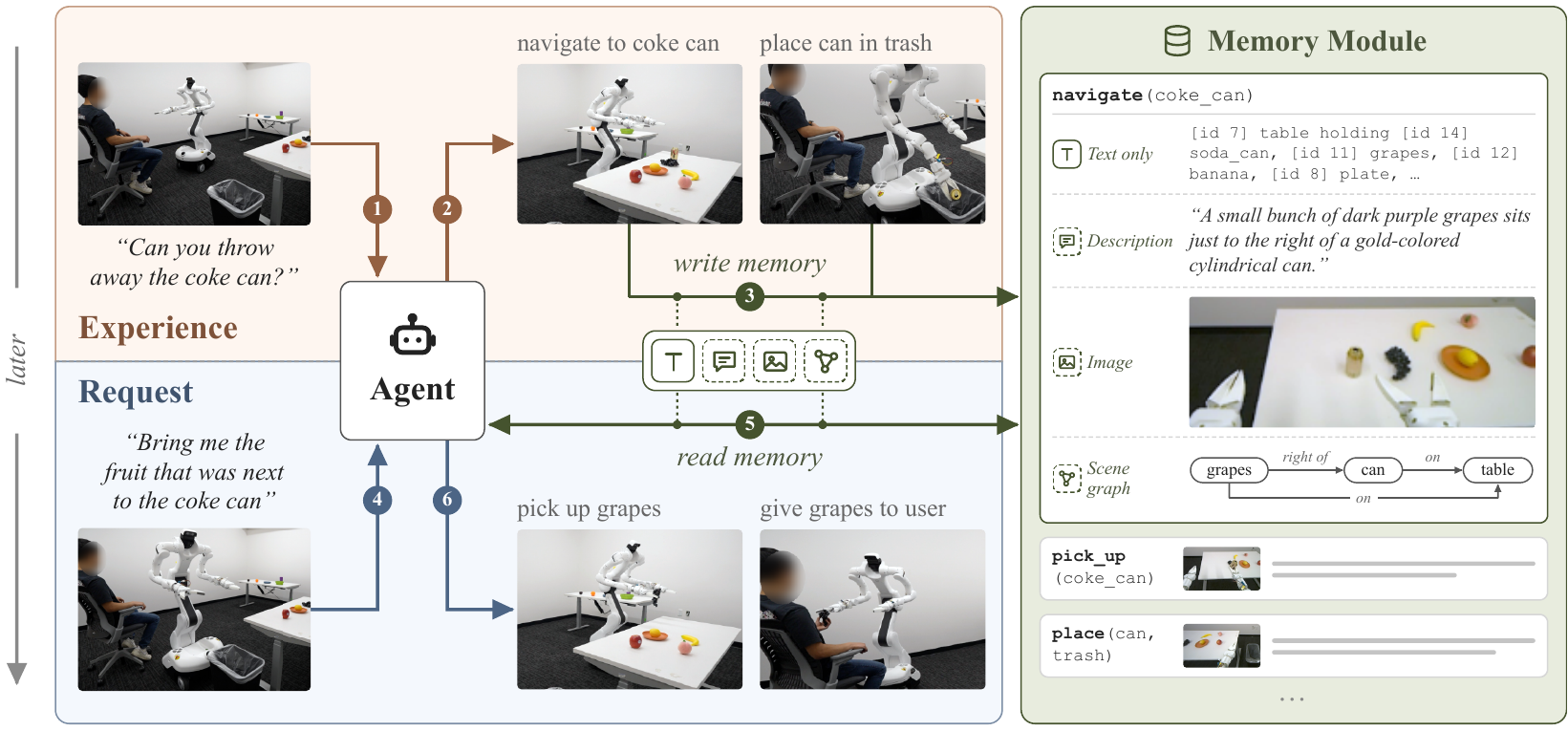}
  \caption{Overview of our embodied agent memory evaluation protocol, on a real robot. During the experience phase, the agent takes in an instruction (1) and executes corresponding actions (2) while writing into its memory (3). During the request phase, the agent receives another instruction (4) and reads its memory (5) for relevant information in order to perform the correct actions (6).}
  \label{fig:teaser}
\end{figure}

\section{Related Work}
\label{sec:related}

\paragraph{Memory for LLM agents.}
Agent memory systems organize retained experience in different ways: MemGPT pages information between the context window and an external archive~\citep{memgpt}, Mem0 extracts and updates facts~\citep{mem0}, A-MEM links evolving notes~\citep{amem}, Reflexion and Agent Workflow Memory distill feedback and reusable procedures~\citep{reflexion,awm}, and filesystem memories let agents read and reorganize files through generic file tools~\citep{fsmemory}. Benchmarks such as LoCoMo and LongMemEval chiefly test recall across long conversations~\citep{locomo,longmemeval}, while MemLens and MemEye evaluate whether memory systems preserve visual evidence across extended multimodal interactions~\citep{memlens,memeye}. We study the corresponding design choice for embodied agents, comparing memory modalities with the agent and observations held fixed and repeating the comparison across memory backends.

\paragraph{Memory for embodied agents.}
Object-centric scene graphs are widely used for robot mapping and planning~\citep{conceptgraphs,karma}, so we include them as one storage modality alongside text, generated descriptions, and frames. Other systems combine scene graphs with images~\citep{grapheqa} or record changing state through text and spatio-temporal relations~\citep{worldlines,daaam,r4dsg}. Retrieval and indexing methods improve the efficiency and accuracy of multimodal memory~\citep{remembr,m3agent,raven} and are complementary to the choice of what to store. Surprise-gated memory addresses which observations to preserve~\citep{surprisemem}; we examine this selection question by comparing writer-selected image retention with archiving every frame.

\paragraph{Embodied benchmarks with memory demands.}
Household benchmarks test long-horizon embodied interaction~\citep{alfred,behavior1k,embodiedbench}, but memory failures there are confounded with exploration, planning, and manipulation ability. Memory-focused evaluations test question answering~\citep{openeqa,mindpalace}, memory-dependent action~\citep{findingdory,memento}, or both~\citep{worldlines}, typically by comparing complete memory systems or agents. When the stored modality does vary, other factors vary with it: in OpenEQA, for example, caption- and scene-graph-based agents use a language model, whereas frame-based agents use a vision-language model~\citep{openeqa}. We design our protocol so that the stored modality is the main variable. Within each backend, the main memory conditions share the same recorded observations, agent, and writer input, and descriptions and frames are archived without passing through the writer. Requests start from a fresh context without new observations, so missing memory cannot be recovered by searching again. No-memory and full-information controls bound performance, information and action requests separate recalling a fact from acting on it, and write-time and read-time inference costs are measured separately.

\section{Preliminaries: LLM-Based Embodied Agents and Memory}
\label{sec:preliminary}

We consider LLM-based embodied agents in which a language or vision-language model selects physical skills for a robot. The interaction loop resembles that of a coding agent: the model receives a task and observations, calls a tool, and uses the returned feedback and updated observations to decide the following action. Each tool corresponds to a physical skill, such as navigating to a room, grasping an object, or opening a drawer, executed by a separate robot controller so the model chooses actions without generating motor commands. This scheme of wrapping robot capabilities as LLM-callable tools is also used by other works~\citep{thea,flexionreflect}. Observations include camera images, detected objects and their locations, user messages, and tool feedback. They may reach the model directly, or information from visual and spatial sensing may be converted into textual records. The observation depends directly on the robot's current state and perception. For example, opening a drawer reveals its contents, while moving to another room removes previously visible objects from view.

We treat memory as an external module that the agent uses through tool calls~\citep{memgpt}. The \emph{agent} is the model together with its current context. At each step it infers the next tool call from its context, and information that leaves the context persists only if the \emph{memory module} retains it. The memory module holds all persistent state and records observations, either through a \emph{writer}, a model that processes observations into memory records, or by archiving artifacts such as camera frames directly. It makes stored information available through context supplied to the agent and retrieval tools for particular records or artifacts, which the agent uses to answer questions or plan actions (Figure~\ref{fig:teaser}). We refer to the agent as the \emph{reader} when it uses stored information in this way. The writer and reader are functional roles and may use the same underlying model, as in our experiments. Some memory configurations require no writer, such as a transcript that stores observations verbatim.

We characterize the memory module by two properties. The \emph{memory modality} is the form of retained evidence, such as textual records, camera frames, or structured spatial records. The \emph{memory backend} is how memory is organized and retrieved, for example as a rolling transcript or interconnected notes. A single backend can support multiple memory modalities, for example by organizing textual entries that reference stored camera frames, generated descriptions, or scene-graph snapshots.

\section{\benchname{}: Evaluation Protocol and Benchmark}
\label{sec:methods}

We introduce \benchname{}, a benchmark for evaluating memory modalities in robotic embodied agents. We first describe its evaluation protocol, including task composition and measured performance metrics, such as task success and inference costs (\S~\ref{sec:tasks}), and then discuss the simulation environments and tasks that instantiate it (\S~\ref{sec:bench}).

\subsection{Evaluation Protocol}
\label{sec:tasks}

To evaluate how stored memory supports later decisions, we divide each task into two phases (Figure~\ref{fig:teaser}): experience, during which the agent observes events and constructs memory, and request, during which it uses that memory to respond to a user without receiving new information.

\textbf{Experience.\;}
In deployment, experience comes from the robot's ordinary operation, such as carrying out an earlier request (Figure~\ref{fig:teaser}). In the benchmark, the robot instead executes a scripted sequence of actions while the memory module records the resulting observations and action outcomes. These actions and surrounding events provide information needed later: the robot may observe a mug being placed inside a cabinet before it is closed, or hear a user identify which of several mugs is theirs. The agent has not yet received any requests, so the memory module must retain information before knowing which details will matter. The experience phase is fixed per task across trials, so differences in performance reflect the memory configuration rather than what was observed.

\textbf{Request.\;}
The request phase begins with a fresh agent context containing no observations from the experience phase. The agent receives a request that requires it to recover some information from its experience through the memory module without receiving new observations. Requests are of two types. \emph{Information requests}, such as asking which cabinet contains a mug, test whether the agent can recover the relevant information from memory. \emph{Action requests}, such as retrieving the mug, test whether the agent can use that information to act. For action requests, the agent submits a complete plan of skill calls, which the robot executes without returning observations or allowing the plan to be revised. This open-loop execution prevents the agent from compensating for missing memory by exploring the scene, for instance by opening cabinets until the mug appears. Within a task, information requests are issued before action requests. Each request starts from a fresh context containing only the request and the name of the robot's current room, which facilitates initial planning without supplying relevant past observations.

\textbf{Stacked tasks.\;}
Separating experience from requests allows tasks to be stacked: the experience phases of several tasks are concatenated into a longer history, followed by all information requests and then all action requests. Before each action request, the simulator is restored to its state at the end of the combined experience phase, so each plan is evaluated from the same physical state without interference from earlier plans. Stacking tests whether earlier information remains available as experience accumulates, approximating the longer histories encountered in deployment. Each component task retains its original grading criteria, allowing the construction of longer-horizon tasks without redesigning individual ones.

\textbf{Outcomes and cost.\;}
Success for an individual task is defined as whether the agent correctly responds to all requests of the task, and for a stacked composite task, it is measured as the fraction of component tasks succeeded. We measure model inference cost separately for memory construction and use. \emph{Write-time cost} accrues during the experience phase as the memory module processes observations, for example by summarizing text or describing images. \emph{Read-time cost} accrues during the request phase as the agent queries memory and produces an answer or plan. Write-time cost is incurred before knowing downstream requests, while read-time cost arises only when responding to requests. Hence, write-time cost scales with the number of actions taken by the agent, and read-time cost scales with the number of requests issued to the agent. For a deployed embodied agent, both costs are highly relevant as robots could take frequent actions in short intervals (e.g., moving through a home through a sequence of navigation skill calls) and could be issued with requests frequently as well. Actual distribution of the two types of cost is determined by the exact application scenario, and therefore we report the two costs separately. To include failed attempts, we also report read-time cost per successful request.

\subsection{Benchmark Environments and Tasks}
\label{sec:bench}

\begin{table}[t]
\caption{Task categories and example evaluation items.}
\label{tab:categories-def}
\centering
\small
\begin{adjustbox}{max width=\linewidth}
\begin{tabular}{@{}lcll@{}}
\toprule
Category & Tasks & Remembered information & Example evaluation items \\
\midrule
Object location       & 11 & where an object was last seen        & ``Which room is the satchel in?'' + fetch \\
Spatial configuration & 11 & metric/ordinal layout                & ``What sat left of the storybook?''; restore a shelf \\
Appearance binding    & 10 & color/pattern tied to an instance    & fetch the \emph{orange} mug among look-alikes \\
Set aggregation       & 8  & counts and collections across rooms  & ``How many mugs are in the kitchen?''; gather all toys \\
User preference       & 5  & user utterance $\to$ instance binding & fetch the mug the user said was theirs \\
Occluded state        & 4  & contents behind doors and lids       & ``What was inside the wardrobe?'' \\
Skill outcome         & 4  & which action variant succeeded       & re-shelve using the placement that worked \\
\bottomrule
\end{tabular}
\end{adjustbox}
\end{table}

\begin{figure}[t]
  \includegraphics[width=\textwidth]{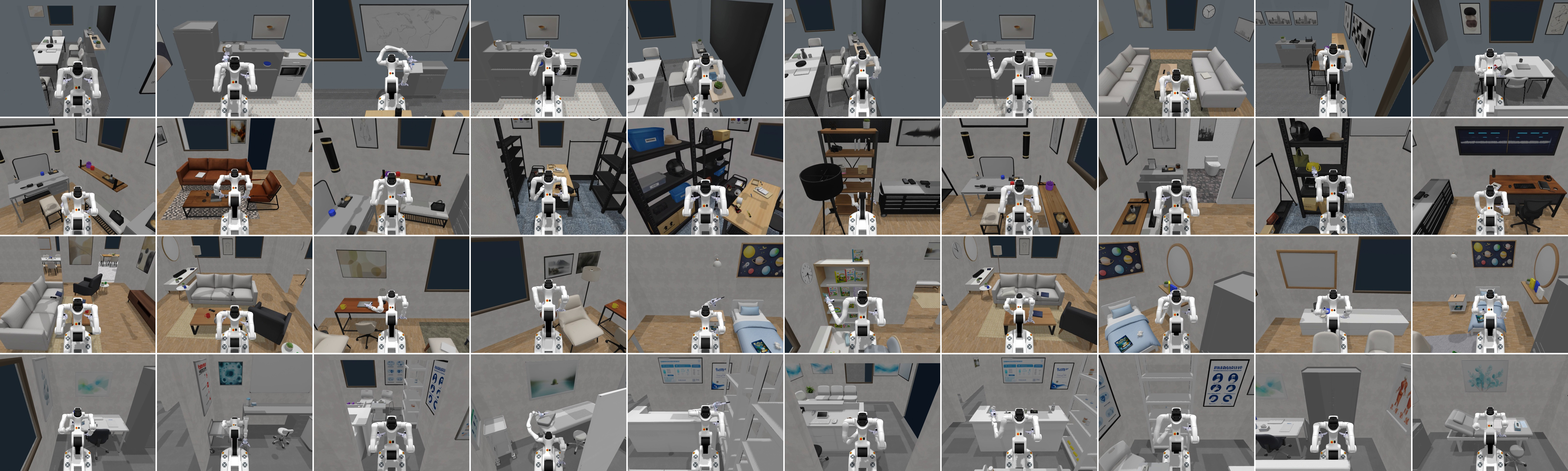}
  \caption{Example snapshots from tasks in \benchname{}. Each row shows different tasks from one scene.}
  \label{fig:tasks}
\end{figure}

\textbf{Environments and robot.\;}
\benchname{} instantiates the protocol in the Genesis simulator~\citep{genesis} with four indoor environments adapted from pre-generated SceneSmith scenes~\citep{scenesmith}: an office, photography studio, home, and clinic, each with seven to nine rooms and 200--300 objects. See Figure~\ref{fig:tasks} for snapshots during example tasks in all four scenes. The simulated robot is a Dexmate Vega 1 bimanual mobile manipulator, which the agent controls through skills that \texttt{navigate} the robot to different places, \texttt{open} and \texttt{close} furniture, and \texttt{pick} and \texttt{place} objects. These skills cover common navigation and manipulation capabilities for deployed robots. Grasping and placement use collision-checked motion plans, and navigation uses planner-validated locomotion paths. During requests, the agent uses \texttt{answer} to respond to information requests and \texttt{submit\_plan} to provide plans for action requests, and signals completion with \texttt{done}.

\textbf{Task categories and construction.\;}
\benchname{} contains 53 tasks across seven categories of remembered information: object location, spatial configuration, appearance binding, set aggregation, user preference, occluded state, and skill outcome (Table~\ref{tab:categories-def}). These categories describe the information that must be remembered, rather than which memory modality is expected to perform best. Each experience phase consists of a scripted tour of the environment, followed by events such as an object being moved, a container being opened, or a user expressing a preference. We also construct a stacked composite task from 10--11 component tasks for each environment. Appendix~\ref{app:setup} provides the individual tasks and composite construction.

\textbf{Grading.\;}
All requests are graded deterministically. Information requests are checked against reference answers of a specified type, such as a region, object, count, or set. Action requests are checked against task-specific predicates on the final simulator state, such as whether the target object is held or rests on the required surface.

\textbf{Observations and representations.\;}
We construct agent observations to reflect information available during robot operation: camera images provide visual evidence, localization and mapping provide object locations and relations, and interactions provide user instructions and action outcomes. The benchmark exposes this information through visual, spatial, and textual observation streams.

Textual observations report user messages, skill feedback, and currently detected objects, including their categories, articulation states, and support and containment relations. They omit color and texture, leaving camera frames as the source of appearance information. Visual observations are camera frames annotated with bounding boxes and integer object identifiers, which the agent also uses to refer to objects in tool calls. Spatial observations are scene-graph snapshots whose nodes store object identifiers and world-frame 3D positions, while edges record support and containment relations~\citep{conceptgraphs,hovsg,supermap}. These snapshots contain no appearance attributes and do not explicitly encode ordinal or viewpoint-dependent relations. In simulation, detections and spatial records are derived from simulator state restricted to observed objects. \S~\ref{sec:realworld} describes a physical instantiation using camera observations and an annotated spatial map.

Given these observations, a memory module can retain textual records, generate image descriptions, archive camera frames, or preserve spatial snapshots. The choice of memory modality determines how the robot's experience remains available for later requests. Description generation and other transformations of observations are part of memory construction. We subsequently describe the configurations evaluated in our experiments in \S~\ref{sec:results-setup}.

\section{Experiments}
\label{sec:results}

\subsection{Experimental Setup}
\label{sec:results-setup}

\textbf{Memory conditions.\;}
Our main comparison varies the evidence archived for later retrieval while holding the memory writer's textual input fixed. For a given backend, task, and seed, the six main conditions share the textual memory constructed from the common observations in \S~\ref{sec:bench}. The \emph{text} condition retains only this memory. The \emph{frames} condition additionally archives the annotated camera frames, while \emph{frames+graph} also archives scene graph snapshots. To test whether textual descriptions of camera frames can replace storing frames themselves, we further design the conditions of \emph{desc@64}, \emph{desc@256}, and \emph{desc@1024}, each archiving generated descriptions of recorded frames at the indicated token budget. Descriptions are generated in separate calls using the same model as the writer and agent, from the annotated frame and the textual observation of the same step. Like frames, they are stored verbatim for the agent to retrieve without passing through the writer. Stored frames, descriptions, and scene-graph snapshots are indexed by the experience step at which they were recorded. Each backend appends this step to every memory entry it renders, and the agent retrieves artifacts by passing step numbers to the retrieval tool of its condition. Frame and description retrieval return the stored artifacts, while scene-graph retrieval returns textual records of the requested objects and their relations. The agent is allowed to make up to five retrieval calls before submitting a response to a request. At most 20{,}000 characters of textual memory are rendered into the agent's context per turn. Retrieved frames, descriptions, and scene-graph records do not count toward this limit.

\textbf{Memory backends.\;}
We evaluate the memory conditions above across five backends that organize memory differently. A \textbf{transcript} retains a rolling history without an LLM-based writer, and a \textbf{wiki} writer maintains named pages through append and rewrite operations. We further adapt three published memory systems to the shared interface and our embodied setting, \textbf{A-MEM}~\citep{amem}, \textbf{Mem0}~\citep{mem0}, and \textbf{ObsMem}~\citep{worldlines}. All writers receive the same observations and instructions on what to remember, and all backends share the memory-context limit and artifact retrieval described above.

\textbf{Reference conditions.\;}
In addition to the five memory backends, we implement two control conditions that bypass memory construction. The \emph{no-memory} condition tests performance without memory, while \emph{full-information} supplies hand-authored textual facts sufficient for each task.

\textbf{Deployment-oriented variants.\;}
The single tasks under the six main conditions form a controlled comparison: experience is short, the writer sees only textual observations, and artifact storage is automatic. Actual deployment of embodied agents could benefit from some relaxations, and we evaluate three of them. First, every main condition also runs on the four stacked composite tasks (\S~\ref{sec:tasks}). Second, because a deployed memory writer would have camera images at hand, \emph{sighted} variants of \emph{frames} and \emph{frames+graph} that also give images to the writer test whether interpreting images during memory construction brings additional benefits over direct archival. Third, because archiving every frame grows storage with the length of deployment, a \emph{selective-storage} variant of \emph{frames} lets the memory writer decide whether to archive each frame using only the textual observations. The writer does not inspect the image, and only retained frames remain retrievable.

\textbf{Models and runs.\;}
The main experiments use Gemini 3.7 Flash. With this model, we evaluate the six main memory conditions across five memory backends on all 53 single tasks and conduct the deployment-oriented evaluations, each across three seeds. To assess whether the relative performance of memory conditions depends on the model, we also evaluate GPT 5.6 Sol and GLM 5.3 Flash on all single tasks, using the six main conditions and three backends (transcript, wiki, and A-MEM) with one seed. Both control conditions are evaluated once per task at a single seed. All models run at temperature 0, and each run uses the same underlying model for the writer, describer, and agent wherever these roles are present.

\textbf{Cost accounting.\;}
We report inference cost in weighted tokens, $T_{\text{in}}\times P_{\text{in}}/P_{\text{out}} + T_{\text{out}}$, where $T_{\text{in}}$ and $T_{\text{out}}$ are the numbers of input and output tokens and $P_{\text{in}}$ and $P_{\text{out}}$ their per-token prices. Input tokens include image tokens and output tokens include reasoning tokens. Weighted tokens therefore equal cost divided by the output-token price, so they are proportional to cost within each model. Because tokenizers and prices differ across models, we compare costs only across conditions within a model. Write-time cost covers writer updates and description generation during experience and is reported per experience step. Read-time cost covers all model calls in the request phase and is reported per attempted request and per successful request.

\subsection{Results}
\label{sec:results-main}

\begin{figure}[tb]
  \includegraphics[width=\textwidth]{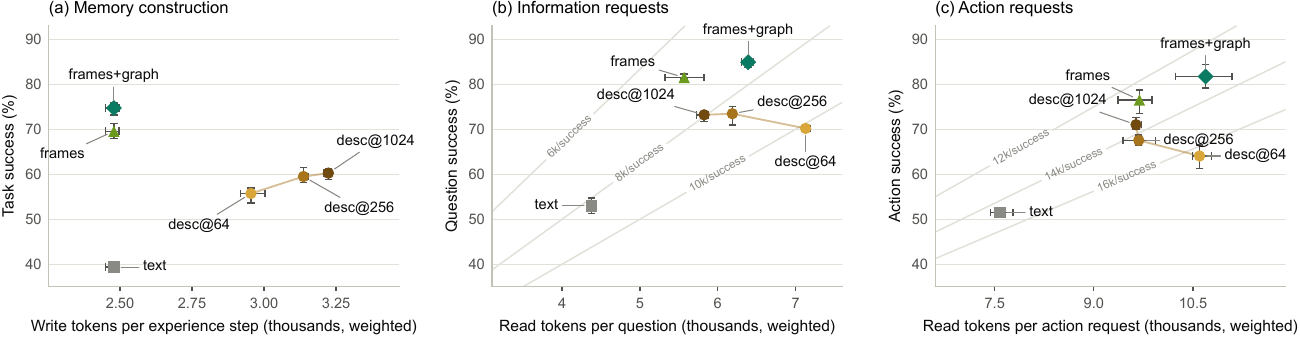}
  \caption{(a) Task success against per-experience-step write cost. (b) Question success against per-information-request read cost. (c) Action success against per-action-request read cost. Aggregated across all tasks, memory backends, and three seeds, for Gemini 3.7 Flash. The transcript backend is excluded from (a) as its writer is not LLM-based. Error bars show the range across seeds.}
  \label{fig:ledger}
\end{figure}

Figure~\ref{fig:ledger} compares task success against memory cost for Gemini 3.7 Flash, averaged across the five memory backends and three seeds. Error bars show the minimum and maximum across seeds. We first examine the effect of memory modality on information and action requests, then its effect on inference costs, and lastly the behavior of different backends under deployment-oriented variants.

\begin{figure}[t]
\centering
\begin{minipage}[t]{0.675\linewidth}
  \centering
  \includegraphics[width=\linewidth]{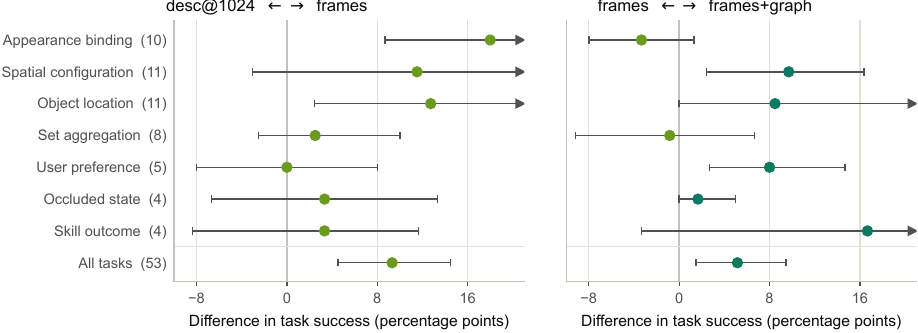}
  \caption{By-category breakdown of differences in task success (\%) between \emph{desc@1024}, frames, and frames with scene graphs. Error bars show 95\% intervals from a paired bootstrap that resamples tasks with replacement.}
  \label{fig:categories}
\end{minipage}\hfill
\begin{minipage}[t]{0.305\linewidth}
  \centering
  \includegraphics[width=\linewidth]{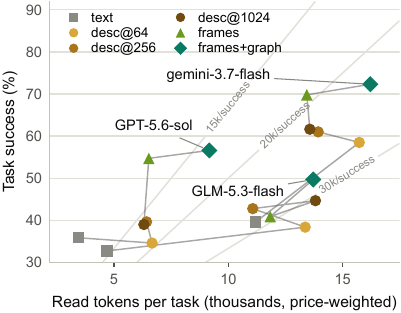}
  \caption{Task success (\%) vs. tokens for three base models, six memory modalities, and three backends on one seed.}
  \label{fig:models}
\end{minipage}
\end{figure}

\textbf{Controls.\;}
Without memory, the agent solves 3 of the 53 tasks; with full-information memory, it solves all 53. Every task is therefore solvable when the relevant facts are explicitly available as text. The following comparisons examine how effectively the tested memory representations preserve evidence that the agent can retrieve and use.

\subsubsection{RQ1: How does memory modality affect task success?}
Storing descriptions of camera frames achieves task success from 56\% to 60\% as the description budget grows from 64 to 1024 tokens, compared against 39\% without any descriptions and 70\% if storing the frames directly (Figure~\ref{fig:ledger}a). Since descriptions and frames are archived independently of the writer and retrieved through the same mechanism, the description gap therefore does not arise from a memory writer further summarizing or omitting the generated descriptions and instead reflects what information the different modalities store and how effectively the reader uses them.

Figure~\ref{fig:categories} shows the paired category-level differences between frames and the longest descriptions, and between frames with and without scene graphs. Error bars show 95\% intervals from a paired bootstrap that resamples tasks with replacement and evaluates both conditions on the same resampled tasks. Frames improve mean success in six of the seven categories, with advantages in appearance binding and object location as shown through the bootstrap interval. Adding scene graphs raises overall success to 75\%, but its effects also vary by category, with advantages on spatial configuration and user preference as reflected by the bootstrap. Storing frames together with scene graphs therefore achieves the highest aggregate task success among the main memory conditions under our implementation.

The limited improvement from increasing the description budget suggests that description length alone does not resolve the challenge of preserving details before their relevance is known. Generating a description commits to an interpretation during memory construction, whereas storing a frame leaves its visual details available for interpretation after the request arrives. The advantage of frames on appearance binding is consistent with this distinction, since identifying a particular object may depend on visual attributes that a description omits.

Scene graphs provide complementary information by making object positions and relations explicit, potentially reducing the need to reconstruct spatial information from camera views. Their benefit on spatial configuration supports this interpretation. The gains are not apparent on set aggregation and occluded state, indicating that the benefit of additional memory modalities varies across task categories.

Figure~\ref{fig:models} extends the comparison to three base models. We primarily compare changes across memory conditions within each model. In terms of task success, storing camera frames and spatial records achieves the highest aggregate task success among the six main conditions for every tested model. Storing frames alone achieves higher task success than \emph{desc@1024} for Gemini 3.7 Flash and GPT 5.6 Sol, but not for GLM 5.3 Flash. Text, frames, and spatial records can encode the same information in different forms: for example, relative object positions may be visible in an image and explicitly recorded in a scene graph. The differing memory-condition rankings suggest that models vary in how effectively they use these representations. Combining textual memory, frames, and scene graphs offers alternative ways to recover shared facts and complementary information, potentially explaining why the combined condition achieves the best performance for each model. The relative benefit of storing frames instead of descriptions therefore depends on the model being used. Differences in read-time token usage further show that models vary in the token expenditure associated with retrieving and using this evidence, yielding distinct trade-offs between task success and token efficiency. We next analyze this trade-off in depth for Gemini 3.7 Flash.

\subsubsection{RQ2: How do memory modalities trade off success against cost?}

\begin{table}[tb]
\caption{Dollar cost and read-time usage by memory condition with Gemini 3.7 Flash, pooling \appMainPool{} and three seeds, at the price of \$0.75 per million input tokens and \$3.75 per million output tokens. Write-time cost is pooled over \appWritePool{} as transcript has no writer. A tool call is either a retrieval, an answer, or a plan.}
\label{tab:app-cost}
\centering
\small
\begin{adjustbox}{max width=\linewidth}
\begin{tabular}{@{}lrrrrr@{}}
\toprule
 & Write \$ per 1k steps $\downarrow$ & \multicolumn{2}{c}{Read \$ per success $\downarrow$} & \multicolumn{2}{c}{Per request} \\
\cmidrule(lr){3-4}\cmidrule(l){5-6}
Condition & & Info. & Action & Calls & Reasoning tokens (k) \\
\midrule
text & 9.30 & 0.031 & 0.055 & 1.0 & 4.2 \\
desc@64 & 11.08 & 0.038 & 0.062 & 2.7 & 4.0 \\
desc@256 & 11.76 & 0.032 & 0.054 & 2.4 & 3.4 \\
desc@1024 & 12.09 & 0.030 & 0.051 & 2.3 & 3.3 \\
frames & 9.30 & 0.026 & 0.047 & 2.1 & 3.1 \\
frames+graph & 9.30 & 0.028 & 0.049 & 2.5 & 3.1 \\
\bottomrule
\end{tabular}

\end{adjustbox}
\end{table}

For Gemini 3.7 Flash, archiving frames, with or without scene graphs, incurs lower write-time inference cost than generating descriptions, which requires additional model calls (Figure~\ref{fig:ledger}a). Table~\ref{tab:app-cost} reports cost and read-time usage across memory conditions, showing that retaining less evidence does not necessarily reduce downstream reasoning. The text condition uses the most reasoning tokens per request despite making the least model calls, suggesting that incomplete evidence can lead to additional deliberation without successful information recovery. Within the description conditions, longer descriptions reduce both model calls and reasoning tokens while providing comparatively small gains in success, despite costing more to generate. This pattern suggests that additional descriptive detail can make retained information easier to use without fully resolving the information needed for the request. Frames use fewer model calls and reasoning tokens than any description condition, consistent with preserved visual evidence reducing the effort required to recover task-relevant information. Adding scene graphs increases model calls while leaving average reasoning-token usage approximately unchanged. Its additional read-time cost is therefore consistent with the expense of retrieving and processing complementary evidence.

The longest descriptions and stored frames incur approximately the same read-time cost per attempted request, but frames achieve higher success. Their lower cost per successful request therefore primarily reflects this success advantage: frames require approximately 15\% fewer weighted tokens per successful information request and 7\% fewer per successful action request than the longest descriptions. Adding scene graphs further improves success but increases read-time cost per successful request. The diagonal guides in Figure~\ref{fig:ledger}b--c show these success-adjusted costs.

\begin{figure}[t]
\centering
\includegraphics[width=\linewidth]{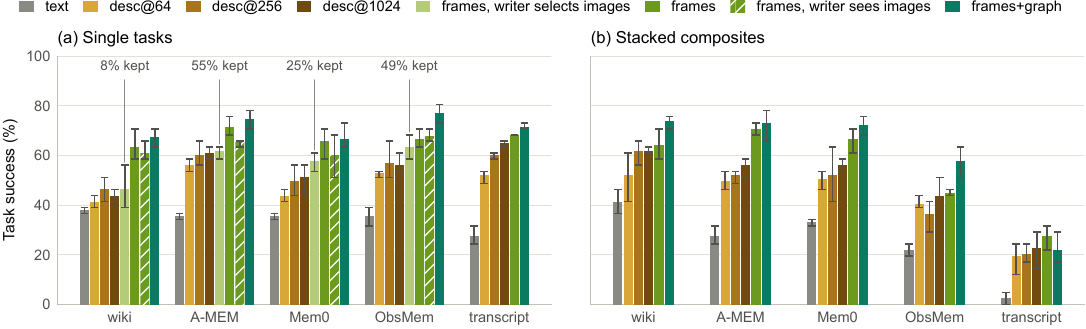}
\caption{Task success (\%) by memory backend and condition with Gemini 3.7 Flash. Both panels evaluate the same component tasks, individually in (a) and after stacking their experience histories in (b). Error bars show the range across seeds. ``Kept'' refers to the percentage of retained frames.}
\label{fig:backends}
\end{figure}

\subsubsection{RQ3: What do deployment-oriented variants show about memory design?}

We next examine the deployment-oriented variants introduced in \S~\ref{sec:results-setup}: accumulated experience, selective image retention, and visual input to the memory writer. These variants extend the main modality comparison to practical memory-design choices while retaining the benchmark's experience-request structure and grading criteria.

\textbf{Accumulated experience.\;}
Figure~\ref{fig:backends} shows that memory backends respond differently to longer histories. Transcript degrades substantially under stacking, because its rolling history exceeds the shared memory-context budget, truncating earlier observations and their retrieval cues from the reader's context. The other adapted backends exhibit different responses to accumulated experience, suggesting that memory organization and retrieval influence how useful evidence remains accessible as histories grow. These results motivate further investigation of the underlying memory design.

\textbf{Selective image retention.\;}
Archiving every observed frame causes storage to grow with experience. Selective retention therefore requires deciding which observations may support requests that are not yet known; recovering omitted evidence may later require additional physical interaction. Under the same selection instruction, the tested writers retain 8-55\% of frames, with lower mean task success than automatic archival for each backend (Figure~\ref{fig:backends}). The differing retention rates and performance changes expose the balance between reducing archival volume and preserving useful evidence, motivating retention policies suited to each backend's memory organization.

\textbf{Visual input to the writer.\;}
Giving the writer access to camera images does not consistently improve task success (Figure~\ref{fig:backends}). Because the reader can already inspect archived frames, visual interpretation by the writer may offer additional value primarily through better memory entries or retrieval cues. The inconsistent success gains suggest that the tested writers do not reliably realize this benefit. Changes in write-time cost further show that visual input affects the memory-construction process even when downstream success remains similar (Appendix~\ref{app:results}).

The changes in relative performance show why memory configurations should be evaluated under the history lengths and retention policies expected in their intended deployment. Together, these variants demonstrate how \benchname{} supports studying memory variants beyond the main modality comparison, exposing design questions relevant to deployed embodied agents.

\subsection{Real-World Instantiation}
\label{sec:realworld}

We deploy our memory evaluation protocol on a physical Dexmate Vega 1 bimanual mobile manipulator, matching the simulated embodiment, using existing controllers for navigation and manipulation. We construct a 3D map using SLAM and manually annotate objects and semantic relations, including support relations, to obtain scene-graph records. Visual observations come from the robot's head-mounted camera, with object segmentation performed using SAM 3~\citep{sam3}.

We design two demonstrations focused on memory of object arrangements. In the first, the robot discards a soda can and should subsequently retrieve the fruit that had been beside it (Figure~\ref{fig:teaser}). In the second, the robot moves a lemon to another table, and should then recall that its original plate was orange, followed by retrieving the apple that had been to the lemon's right.

In the two demonstrations, the robot correctly used its memory and completed both deliveries without human assistance. We used \emph{wiki} as the memory backend, using the \emph{frames+graph} memory modality. These demonstrations establish that the protocol and memory interfaces can be instantiated on a physical robot using camera observations and a manually annotated spatial map, providing a concrete basis for future comparisons of memory configurations in physical environments.

\section{Conclusion}
\label{sec:conclusion}

We present \benchname{}, a benchmark for evaluating memory in embodied agents through a controlled experience–request protocol, which tests whether retained observations support subsequent recall and action, while measuring inference costs separately for memory construction and use. Our reference evaluations show that memory modality affects both task success and inference cost, with benefits varying across models and task categories. For the primary model, retaining camera frames achieves higher success than generating descriptions at lower write-time inference cost, while adding scene graphs further improves success at higher read-time cost.

Deployment-oriented variants demonstrate how the benchmark supports investigating additional memory-design choices. Accumulated experience affects backends differently, and selecting frames from textual observations reduces archival volume with losses in task success. Giving images to the writer does not consistently improve success, although it can reduce write-time cost for some backends. These findings motivate studying how memory organization, retention decisions, and writer inputs interact. Two physical-robot demonstrations instantiate the protocol using camera observations and an annotated spatial map, showing that the benchmark’s protocol design can support evaluations beyond simulation. By providing common experiences, interfaces, and grading, \benchname{} offers an extensible experimental basis for developing memory systems for embodied agents.

\bibliography{references}
\bibliographystyle{plainnat}

\appendix
\clearpage
\section{Additional Results}
\label{app:results}

Tables~\ref{tab:app-main}, \ref{tab:app-categories}, \ref{tab:app-models}, and~\ref{tab:app-backends} list the values behind Figures~\ref{fig:ledger}--\ref{fig:backends}, and Table~\ref{tab:app-writer} gives the cost of writer image access by backend.

\begin{table}[ht]
\caption{Values behind Figure~\ref{fig:ledger}: success and inference cost by memory condition with Gemini 3.7 Flash, pooling 53 tasks, \appMainPool{}, and three seeds (\appMainRuns{} runs per condition). Info.\ and Action give success on information and action requests. Write-time tokens are pooled from \appWritePool{} as transcript has no writer. The reference conditions run once per task at seed 0. Per-seed task success varies by at most \mainSeedSpread{} percentage points within a condition.}
\label{tab:app-main}
\centering
\small
\begin{adjustbox}{max width=\linewidth}
\begin{tabular}{@{}lrrrrrr@{}}
\toprule
 & \multicolumn{3}{c}{Success (\%) $\uparrow$} & Write tokens per step $\downarrow$ & \multicolumn{2}{c}{Read tokens per success $\downarrow$} \\
\cmidrule(lr){2-4}\cmidrule(l){6-7}
Condition & Task & Info. & Action & (k, weighted) & Info. (k) & Action (k) \\
\midrule
text & 39 & 53 & 51 & 2.48 & 8.3 & 14.7 \\
desc@64 & 56 & 70 & 64 & 2.95 & 10.2 & 16.5 \\
desc@256 & 59 & 73 & 68 & 3.14 & 8.4 & 14.3 \\
desc@1024 & 60 & 73 & 71 & 3.22 & 8.0 & 13.6 \\
frames & 70 & 82 & 77 & 2.48 & 6.8 & 12.7 \\
frames+graph & 75 & 85 & 82 & 2.48 & 7.5 & 13.1 \\
\midrule
no memory & 6 & 17 & 0 & -- & -- & -- \\
full information & 100 & 100 & 100 & -- & -- & -- \\
\bottomrule
\end{tabular}

\end{adjustbox}
\end{table}

\begin{table}[ht]
\caption{Values behind Figure~\ref{fig:categories}: task success (\%) by category with Gemini 3.7 Flash, pooling \appMainPool{} and three seeds, and the plotted differences with 95\% intervals from a paired bootstrap that resamples tasks (\appBootN{} resamples). Each task contributes \appRunsPerTask{} runs per condition.}
\label{tab:app-categories}
\centering
\small
\begin{adjustbox}{max width=\linewidth}
\begin{tabular}{@{}lcccccc@{}}
\toprule
 & & \multicolumn{3}{c}{Success (\%)} & \multicolumn{2}{c}{Difference (pp) [95\% interval]} \\
\cmidrule(lr){3-5}\cmidrule(l){6-7}
Category & Tasks & desc@1024 & frames & frames+graph & \makecell{frames\\$-$ desc@1024} & \makecell{frames+graph\\$-$ frames} \\
\midrule
Appearance binding & 10 & 55 & 73 & 69 & $+18.0$ [$8.7$, $27.3$] & $-3.3$ [$-8.0$, $1.3$] \\
Spatial configuration & 11 & 56 & 67 & 77 & $+11.5$ [$-3.0$, $27.9$] & $+9.7$ [$2.4$, $16.4$] \\
Object location & 11 & 77 & 90 & 98 & $+12.7$ [$2.4$, $24.2$] & $+8.5$ [$0.0$, $22.4$] \\
Set aggregation & 8 & 58 & 61 & 60 & $+2.5$ [$-2.5$, $10.0$] & $-0.8$ [$-9.2$, $6.7$] \\
User preference & 5 & 63 & 63 & 71 & $+0.0$ [$-8.0$, $8.0$] & $+8.0$ [$2.7$, $14.7$] \\
Occluded state & 4 & 48 & 52 & 53 & $+3.3$ [$-6.7$, $13.3$] & $+1.7$ [$0.0$, $5.0$] \\
Skill outcome & 4 & 53 & 57 & 73 & $+3.3$ [$-8.3$, $11.7$] & $+16.7$ [$-3.3$, $36.7$] \\
\midrule
All tasks & 53 & 60 & 70 & 75 & $+9.3$ [$4.5$, $14.5$] & $+5.2$ [$1.5$, $9.4$] \\
\bottomrule
\end{tabular}

\end{adjustbox}
\end{table}

\begin{table}[ht]
\caption{Values behind Figure~\ref{fig:models}: task success and read tokens by base model at seed 0, pooling three backends (159 runs per cell). Each model serves as writer, describer, and agent on the same recorded tours. Weighted tokens are $T_{\text{in}} \times P_{\text{in}}/P_{\text{out}} + T_{\text{out}}$ per task, at the prices of \$0.75/\$3.75 per million input/output tokens for Gemini 3.7 Flash, \$4/\$20 for GPT 5.6 Sol, and \$0.15/\$0.5 for GLM 5.3 Flash.}
\label{tab:app-models}
\centering
\small
\begin{adjustbox}{max width=\linewidth}
\begin{tabular}{@{}lcccccc@{}}
\toprule
Model & text & desc@64 & desc@256 & desc@1024 & frames & frames+graph \\
\midrule
\multicolumn{7}{@{}l}{\emph{Task success (\%)}} \\
Gemini 3.7 Flash & 40 & 58 & 61 & 62 & 70 & 72 \\
GPT 5.6 Sol & 36 & 35 & 40 & 39 & 55 & 57 \\
GLM 5.3 Flash & 33 & 38 & 43 & 45 & 41 & 50 \\
\midrule
\multicolumn{7}{@{}l}{\emph{Read tokens per task (thousands, price-weighted)}} \\
Gemini 3.7 Flash & 11.2 & 15.7 & 13.9 & 13.6 & 13.4 & 16.2 \\
GPT 5.6 Sol & 3.4 & 6.7 & 6.4 & 6.3 & 6.5 & 9.2 \\
GLM 5.3 Flash & 4.7 & 13.4 & 11.1 & 13.8 & 11.8 & 13.7 \\
\bottomrule
\end{tabular}

\end{adjustbox}
\end{table}

\begin{table}[ht]
\caption{Values behind Figure~\ref{fig:backends}: task success (\%) by memory backend and condition with Gemini 3.7 Flash, pooling three seeds. Single-task rows cover the 41 tasks that appear in the stacked composites. In a stacked composite, a component task passes when all of its requests pass. In the writer-selects condition, the writer chooses which frames to archive; kept is the percentage of offered frames it retains, with one frame offered per experience step across the scene tours and task experience segments. The transcript has no writer. All backends are our adaptations.}
\label{tab:app-backends}
\centering
\small
\begin{adjustbox}{max width=\linewidth}
\begin{tabular}{@{}lcccccccccc@{}}
\toprule
 & \multicolumn{6}{c}{Main conditions} & \multicolumn{3}{c}{frames variants} \\
\cmidrule(lr){2-7}\cmidrule(l){8-10}
Backend & text & desc@64 & desc@256 & desc@1024 & frames & frames+graph & \makecell{writer\\selects} & \makecell{kept\\(\%)} & \makecell{writer sees\\images} \\
\midrule
\multicolumn{10}{@{}l}{\emph{Single tasks}} \\
wiki & 38 & 41 & 46 & 44 & 63 & 67 & 46 & 8 & 61 \\
A-MEM & 36 & 56 & 60 & 61 & 72 & 75 & 62 & 55 & 65 \\
Mem0 & 36 & 44 & 50 & 51 & 66 & 67 & 58 & 25 & 60 \\
ObsMem & 36 & 53 & 57 & 56 & 67 & 77 & 63 & 49 & 67 \\
transcript & 28 & 52 & 60 & 65 & 68 & 72 & -- & -- & -- \\
\midrule
\multicolumn{10}{@{}l}{\emph{Stacked composites (component tasks passed)}} \\
wiki & 41 & 52 & 62 & 62 & 64 & 74 & -- & -- & -- \\
A-MEM & 28 & 50 & 52 & 56 & 71 & 73 & -- & -- & -- \\
Mem0 & 33 & 50 & 52 & 56 & 67 & 72 & -- & -- & -- \\
ObsMem & 22 & 41 & 37 & 44 & 45 & 58 & -- & -- & -- \\
transcript & 2 & 20 & 20 & 23 & 28 & 22 & -- & -- & -- \\
\bottomrule
\end{tabular}

\end{adjustbox}
\end{table}

\begin{table}[ht]
\caption{Task success and cost of writer image access by backend with Gemini 3.7 Flash, on all 53 tasks and three seeds. The default writer receives only textual observations, with the alternative condition being it also sees camera frames. The transcript has no writer and is omitted.}
\label{tab:app-writer}
\centering
\small
\begin{adjustbox}{max width=\linewidth}
\begin{tabular}{@{}lcccccc@{}}
\toprule
 & \multicolumn{2}{c}{Task success (\%)} & \multicolumn{2}{c}{Write tokens per step (k)} & \multicolumn{2}{c}{Read tokens per task (k)} \\
\cmidrule(lr){2-3}\cmidrule(lr){4-5}\cmidrule(l){6-7}
Backend & default & \makecell{writer sees\\images} & default & \makecell{writer sees\\images} & default & \makecell{writer sees\\images} \\
\midrule
\multicolumn{7}{@{}l}{\emph{frames}} \\
wiki & 67 & 64 & 1.48 & 1.81 & 14.1 & 14.6 \\
A-MEM & 69 & 64 & 1.73 & 1.90 & 11.3 & 11.6 \\
Mem0 & 69 & 60 & 3.22 & 2.68 & 13.9 & 14.9 \\
ObsMem & 68 & 68 & 3.48 & 2.08 & 14.8 & 15.1 \\
\midrule
\multicolumn{7}{@{}l}{\emph{frames+graph}} \\
wiki & 70 & 73 & 1.48 & 1.89 & 15.6 & 17.5 \\
A-MEM & 74 & 71 & 1.73 & 1.88 & 13.3 & 13.4 \\
Mem0 & 72 & 72 & 3.22 & 2.67 & 15.1 & 16.0 \\
ObsMem & 80 & 77 & 3.48 & 2.09 & 16.6 & 16.7 \\
\bottomrule
\end{tabular}

\end{adjustbox}
\end{table}

\clearpage
\section{Task inventory}
\label{app:setup}

\begin{table}[ht]
\caption{The 53 tasks by scene and category: loc (object location), spat (spatial configuration), app (appearance binding), set (set aggregation), pref (user preference), occl (occluded state), and skill (skill outcome). Unless marked, a task has one information request and one action request. Tasks marked info only have no action request, skill-outcome tasks have only an action request, and three preference tasks and qa\_inventory have two information requests. Tasks marked $\dagger$ are not members of the scene's stacked composite.}
\label{tab:app-inventory}
\centering
\small
\begin{adjustbox}{max width=\linewidth}
\begin{tabular}{@{}l>{\raggedright\arraybackslash}p{0.8\linewidth}@{}}
\toprule
Scene & Tasks (category) \\
\midrule
office (12) & loc\_kettle (loc), reloc\_tumbler (loc), vis\_book\_room (loc), spatial\_row\_neighbor (spat), vis\_tumbler\_color (app), vis\_mug\_perch (app), vis\_bowl\_stack (app), vis\_bowl\_station (app), qa\_inventory$^\dagger$ (set, info only), collect\_succulents (set), pref\_water\_carafe (pref), proc\_credenza\_return$^\dagger$ (skill) \\
studio (14) & loc\_gaffer\_tape (loc), reloc\_satchel (loc), vis\_mug\_room (loc), spatial\_shelf\_vacate (spat), vis\_row\_neighbor (spat), vis\_stack\_order$^\dagger$ (spat), vis\_cart\_clearout (spat), vis\_candle\_color (app), qa\_office\_twins$^\dagger$ (app, info only), qa\_inventory\_candles$^\dagger$ (set, info only), collect\_shoes (set), pref\_client\_plant (pref), vis\_drawer\_reveal (occl), proc\_shelf\_return$^\dagger$ (skill) \\
home (14) & loc\_teakettle (loc), reloc\_remote (loc), vis\_teddy\_rooms (loc), spatial\_row\_neighbor (spat), spatial\_shelf\_vacate (spat), vis\_stack\_order$^\dagger$ (spat), vis\_shelf\_clearout (spat), vis\_mug\_color (app), vis\_storybook\_cover$^\dagger$ (app, info only), qa\_inventory\_mugs$^\dagger$ (set, info only), collect\_toys (set), pref\_snack (pref), vis\_container\_reveal (occl), proc\_desk\_return$^\dagger$ (skill) \\
clinic (13) & loc\_diagnostic\_kit (loc), reloc\_clipboard (loc), spatial\_row\_neighbor (spat), spatial\_shelf\_vacate (spat), vis\_sanitizer\_color (app), vis\_medbox\_shelf (app), qa\_inventory\_tissues (set, info only), collect\_gloveboxes (set), pref\_allergy\_med (pref), pref\_exam\_setup (pref), vis\_cabinet\_reveal (occl), vis\_locker\_reveal$^\dagger$ (occl, info only), proc\_pharmacy\_return$^\dagger$ (skill) \\
\bottomrule
\end{tabular}
\end{adjustbox}
\end{table}

\end{document}